\documentclass{svproc}
\usepackage{url}

\usepackage{graphicx}
\usepackage{subfig}
\usepackage{mathtools}

\begin{document}
\mainmatter              
\title{Double Deep Reinforcement Learning Techniques for Low Dimensional Sensing Mapless Navigation of Terrestrial Mobile Robots}
\titlerunning{Double Deep Reinforcement Learning}  
%
\author{Linda Dotto de Moraes\inst{1} \and
Victor Augusto Kich\inst{1} \and Alisson Henrique Kolling\inst{1} \and
Jair Augusto Bottega\inst{1} \and Raul Steinmetz\inst{1} \and Emerson Cassiano da Silva\inst{1} \and Ricardo Grando\inst{2} \and Anselmo Rafael Cuckla\inst{1} \and Daniel Fernando Tello Gamarra\inst{1}}
\authorrunning{Linda Dotto de Moraes et al.} 
%
\tocauthor{Ivar Ekeland, Roger Temam, Jeffrey Dean, David Grove,
Craig Chambers, Kim B. Bruce, and Elisa Bertino}

\institute{Universidade Federal de Santa Maria, Santa Maria, Brazil 
\email{lindadotto@yahoo.com.br}\\ \and
Universidad Tecnologica del Uruguay, Uruguay
\email{ricardo.bedin@utec.edu.uy}\\
}

\maketitle              

\begin{abstract}
In this work, we present two Deep Reinforcement Learning (Deep-RL) approaches to enhance the problem of mapless navigation for a terrestrial mobile robot. Our methodology focus on comparing a Deep-RL technique based on the Deep Q-Network (DQN) algorithm with a second one based on the Double Deep Q-Network (DDQN) algorithm. We use 24 laser measurement samples and the relative position and angle of the agent to the target as information for our agents, which provide the actions as velocities for our robot. By using a low-dimensional sensing structure of learning, we show that it is possible to train an agent to perform navigation-related tasks and obstacle avoidance without using complex sensing information. The proposed methodology was successfully used in three distinct simulated environments. Overall, it was shown that Double Deep structures further enhance the problem for the navigation of mobile robots when compared to the ones with simple Q structures.

\end{abstract}
\section{Introduction}
The field of robotics has an intrinsic relation with Reinforcement Learning (RL) since many problems in robotics can use this artificial intelligence technique. Many algorithms have already been developed for a range of tasks to teach an agent to learn a perfect set of actions through trial-and-error interactions. The feedback that the agent receives from the workplace guides the learning step-by-step toward an optimal policy \cite{kober2013reinforcement}. 
These techniques have achieved a state-of-art performance in some problems in robot learning, given the evolution of deep learning neural networks (Deep ANN). By representing the agent as a Deep ANN, it was possible to improve the agent's ability to evolve around complex environments and with distinct tasks. However, since this improvement relies on ANN constraints, such as the gradient for learning, other problems have emerged - namely, the problem related to the ability to learn from high-dimensional data. Given this limitation, specific Deep-RL techniques, such as the Contrastive Learning \cite{laskin2020curl} have been used to suppress and make the agent learn. For navigation-related tasks in specific, it has been shown that a good performance is possible to be yielded through simple sensing information, like for terrestrial mobile robots \cite{tai2017virtual} \cite{jesus2021soft}, aerial robots \cite{grando2022double}, underwater robots \cite{carlucho2018}, and even hybrid robots \cite{bedin2021deep}.
 This so-called mapless navigation is the base for many problems in mobile robotics, where many Deep-RL algorithms have been employed.

\begin{figure}[bp!]
    \centering
    \includegraphics[width=0.6\linewidth]{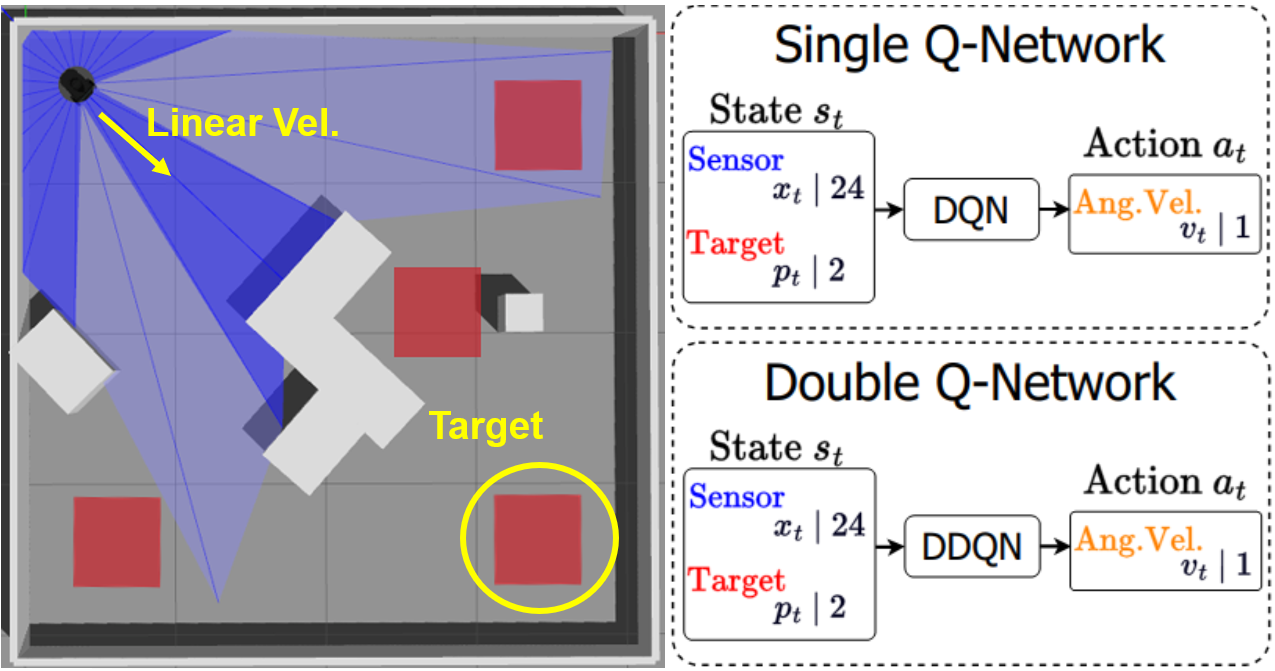}
    \caption{Turtlebot3 Burger performs among obstacles in one of the scenarios (left) and the model structure with inputs and outputs of our methods, DQN and Double DQN (right).}
    \label{fig:structure}
\end{figure}

With that in mind, this research proposes to demonstrate and evaluate two Deep-RL techniques' effectiveness in tasks involving the goal-oriented navigation of a terrestrial mobile robot. The DQN and DDQN algorithms are used, performing a comparison. The approaches are based on the simple sensing idea, with an architecture of 26 samples for the state, arranged by 24 laser sensor readings and the distance and angle of the mobile robot related to the target. We also focus on showing how Double Q architectures perform when compared with simple Q architectures. Figure \ref{fig:structure} shows our proposed architecture for learning. 

Overall, this work has the following contributions: 1) It is shown that Double Q approaches are capable of performing mapless navigation of terrestrial mobile robots; 2) It is shown that a combination of Double Q approaches and low dimensional sensing deal better with some emergent problems in Deep-RL, such as the convergence of the gradient descent or the catastrophic forgetting problems, consistently softening them;

This work has seven sections. After a brief introduction, the works of other researchers in the field that served as inspiration for this one are described in Section \ref{related_works}. Section \ref{theoretical_background} provides a theoretical basis for the algorithms
used in the experiments. In Section \ref{expreimental_setup}, the tools, software, and environments used in this paper are described. The methodology used to teach the agent how to reach a target is shown in Section \ref{methodology}, along with an explanation of the network structure and reward function. The results achieved in this study are detailed in Section \ref{results}.
Finally, the last section discusses the main
results achieved and applications of Deep-RL.

\section{Related Works}\label{related_works}

Deep-RL has been applied previously to mapless navigation tasks for mobile robots
\cite{tai2017virtual},
\cite{chen2017socially}, \cite{jesus2019deep}, \cite{jesus2021soft}.
Mnih et al. \cite{mnih2013playing} calculated a value function for future reward using an algorithm called deep Q-network (DQN) for the Atari games, \cite{hausknecht2015deep}, \cite{mnih2013playing}, \cite{van2016deep}.
It is necessary to outline that the DQN just uses discrete actions when it is applied to a problem like the control of a robot.
In order to extend  the DQN to a continuous control, Lillicrap et al. \cite{lillicrap2015continuous} proposed a deep deterministic policy gradients (DDPG) algorithm.
This algorithm paves the way to the use of Deep-RL in mobile robot navigation.

Dobrevski et al. \cite{dobrevski2018map} have adopted an approach for mapless goal-driven navigation optimizing a behavioral policy within the framework of reinforcement learning, specifically the Advantage Actor-Critic (A2C) method. The results are compared with the performance of the standard Turtlebot3 navigation package and it was found that the approach reached a more robust performance.

Tai et al. \cite{tai2017virtual} developed a mapless motion planner for a mobile robot and used as input to their system the range findings sensor information and the target position and the output was the continuous steering commands for the robot. Initially, they employed discrete steering commands on \cite{tai2016towards}. It was demonstrated that, with the asynchronous Deep-RL method, it is possible to train an agent to arrive to a predetermined target using a mapless motion planner. 

As in the work of Tai in \cite{tai2017virtual} and the others related, this paper focuses on developing a mapless motion planner based on low-dimensional range readings. Our work differs by using a deterministic approach based on the DDQN for the resolution of navigation-related problems for terrestrial mobile robots and the insertion of a dynamic target for the robot in the environments without an asynchronous training. Overall, we show that these tasks for terrestrial mobile robots can be excelled using low dimensional sensing data and simple Deep-RL approaches, such as de DDQN. By this mean, we show that a common problem in Deep-RL such as the convergence of the gradient descent or the catastrophic forgetting problems can be consistently smoothed.

\section{Theorical Background}\label{theoretical_background}

\subsection{Deep Q-Network - DQN}

At the forefront of the recent Deep-RL breakthroughs, we have the Deep-Q Network(DQN)\cite{mnih2013playing} method, developed by Mnih et al.
Applying the concepts of reinforcement learning, such as the \textit{Bellman equation} given by:

\begin{equation}
    Q^{*}(s,a)=E_{s'\sim\varepsilon}\left[r + \gamma \underset{a'}{max}Q^*(s',a')|s,a\right]
    \label{eq:bellman}
\end{equation}

This equation returns the optimal value given a state-action pair. Employing a neural network with weights $\theta$ as a function approximator to estimate the action-value function in the \textit{Bellman equation}, we have $Q(s,a,\theta) \approx Q^*(s,a)$, guaranteeing the convergence to the optimal value. This neural network can be trained by minimizing a loss $\mathcal{L}(\theta)$ given by the following equation:
\begin{equation}\label{eq:loss}
    \mathcal{L}(\theta) = [(y - Q(s,a,\theta))^2].
\end{equation}
With $y=E_{s'\sim\varepsilon}[r + \gamma \underset{a'}{max}Q(s',a';\theta_{t})|s,a]$ being a target function, employing the network weights from the $t$ episode, used to create a temporal difference between the value found and the expected value.

The inputs to the neural network are a state-action pair sampled randomly from an experience replay buffer, which stores each transition achieved by the agent. This makes the method \textit{model-free}, reducing the problems of correlated data and non-stationary distributions. The action chosen by the agent follows a $\epsilon$-greedy policy, with $a = \underset{a}{max}Q(s,a;\theta)$ when $1 - \epsilon$ is true or a random action with $\epsilon$ probability.

\subsection{Double Deep-Q Network - DDQN}

In regular Q-Learning and Deep Q-Network, the max operator employs the same values for selecting and evaluating an action. This increases the likelihood of choosing overestimated values, leading to overly optimistic value estimates. As a potential solution for the overestimated values issue, Hasselt \cite{hasselt2010double} developed a new algorithm called Double Q-leaning, which uses a double estimator approach to estimate the value of the next state. 
Using the Double Q-Learning strategy associated with DQN, Van Hasselt et al. \cite{van2016deep} bring a new approach called Double Deep-Q Network (DDQN).
In DDQN, the target function is presented as 
\begin{equation}
y_{t}^{DoubleDQN}  = r_{t+1} + \gamma Q(S_{t+1}, \underset{a}{argmax}Q(s_{t+1},a;\theta_{t}),\theta_{t}^{-}),
\end{equation}
where $\mathcal{L}(\theta_{t}^{-})$ is the weight of the target network for the evaluation of the current greedy policy.

\section{Experimental Setup}\label{expreimental_setup}

In this section we describe the robots, tools, and environments used in the project.

\subsection{PyTorch}

Python was the programming language used in the development of the algorithms, and it has applications in different areas, such as image processing\cite{Pfitscher2019} and robotics  \cite{daSilva2019}. 
One of Python's libraries is PyTorch. PyTorch is an open-source machine learning library applied here on the creation of deep neural networks \cite{subramanian2018deep}. It provides two high-level features: tensor computation with acceleration via a graphical processing unit and a tape-based automatic differentiation system. 

\subsection{ROS, Gazebo and Turtlebot}
The robot operating system (ROS) is an open-source robotics middleware that acts as an intermediary, managing the connections between software and hardware in robots.
ROS \cite{pyo2015ros} is not an operating system but provides some operational system standard services, like messages between processes, low-level device control, package management and hardware abstraction.  
Graph architecture is utilized to represent the processes running by ROS, where the processes are called nodes and messages between two or more processes communicating are the topics. For message exchange between nodes and topics, the concept of publishing and subscribing is used. 

As a support tool for simulated experiments, the open-source 3D simulator Gazebo\cite{fairchild2016ros} is an excellent facilitator for developments using ROS. The simulation of different environments is one of the biggest advantages of its integration with ROS because Gazebo has real-world rules and concepts, facilitating the simulation of applicability. 

The mobile robot used is the Turtlebot 3, from Robotis company.
Because it is a robot with low hardware costs and open source software, it facilitates the implementation of projects. The version used in the experiments was TurtleBot3, with Raspberry Pi3 development hardware, which further enhances the vast community that uses it.

\subsection{Simulation Environments}

\begin{figure}[bp!]
  \subfloat[First scenario.\label{fig:env_1_sim}]{
	\begin{minipage}[c][\width]{0.24\textwidth}
	   \centering
	   \includegraphics[width=\textwidth]{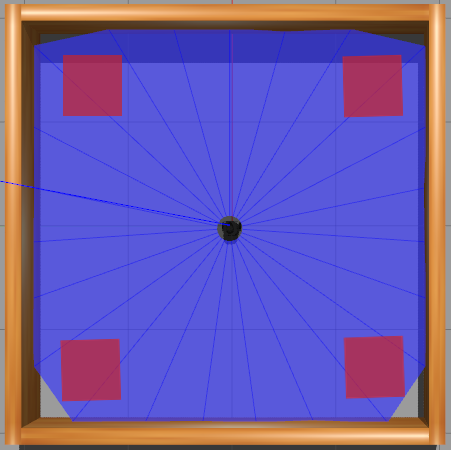}
	\end{minipage}}
 \hfill 	
  \subfloat[Second scenario.\label{fig:env_2_sim}]{
	\begin{minipage}[c][\width]{0.24\textwidth}
	   \centering
	   \includegraphics[width=\textwidth]{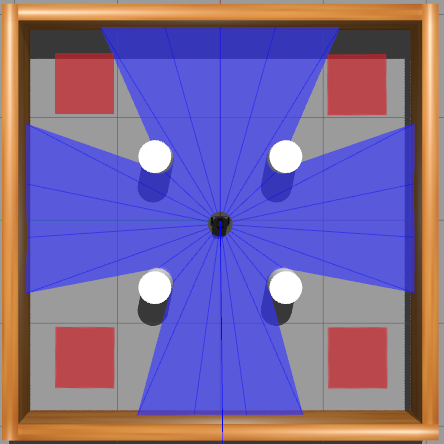}
	\end{minipage}}
 \hfill 
 \subfloat[Third scenario.\label{fig:env_3_sim}]{
	\begin{minipage}[c][\width]{0.24\textwidth}
	   \centering
	   \includegraphics[width=\textwidth]{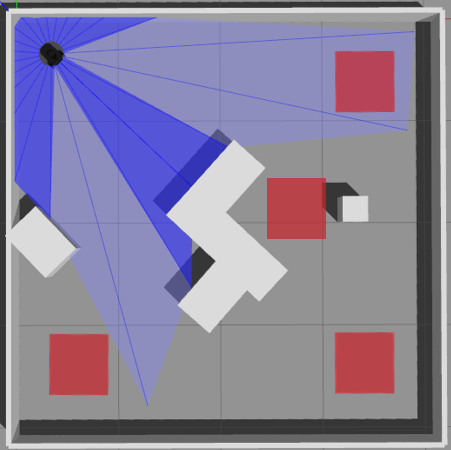}
	\end{minipage}}
\caption{Simulated scenarios created in the Gazebo simulator.}
\label{fig:environments_sim}
\end{figure}

Three simulated environments were constructed in Gazebo, which are limited by walls. The first one is empty and without obstacles besides the walls, and it can be observed in Figure \ref{fig:env_1_sim}. The image also shows the position of the four fixed goals used on the test phase.
Aiming to add challenge to the next scenario, in the second environment, four cylinders were placed as fixed obstacles, as shown in Figure \ref{fig:env_2_sim}.
In the third environment, the obstacles are white blocks placed in the middle of the scenario. And as can be seen in Figure \ref{fig:env_3_sim}, in contrast to the first and second environments, in the third environment, the robot starts the simulation positioned near the corner.
The agent receives a negative reward each time it has a collision with a wall or an obstacle, and the episode ends. If the robot reaches the target, the episode also ends.

In the training phase, the target is placed in a new random position on the scenario map each time a new episode begins.
By doing this, the scenario becomes more dynamic, and the robot is prevented from memorizing a navigation strategy.
During the simulated testing phase, the target was placed in four fixed positions to ensure that the comparison between the results of the two algorithms in the same environment could be more accurate.

\section{Methodology}\label{methodology}
The goal of this work is to train, test, and compare the performance of two algorithms, DQN and Double DQN, applied to a mobile robot. The algorithms aim to support the robot in order to avoid walls and obstacles, planning its movements to navigate without map knowledge on the environment.
The robot is set with a fixed linear velocity and varies between five angular velocities, which are the five actions that the robot can take according to the network output.

\subsection{Network Structure}
\begin{figure}
    \centering
    \includegraphics[width=0.29\linewidth]{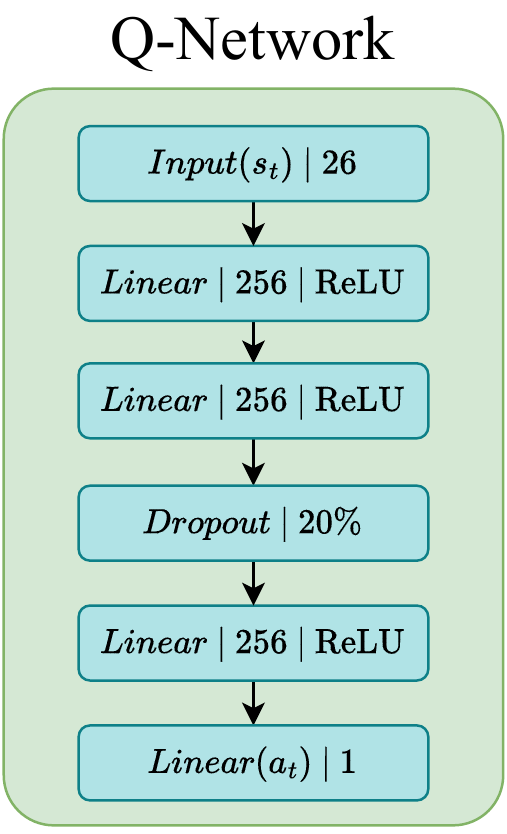}
    \caption{Q-Network architecture used in both DQN and DDQN.}
    \label{fig:architecture}
\end{figure}

After the system's states and actions were established, a Q-Network was created to compose the DQN and Double DQN architectures.
The Q-network has 26 inputs, including the laser range sensor measurements, previous angular and linear velocities, and the position and orientation of the target. The output of the network is a discrete value between 0 and 4 and is equivalent to the angular velocity, as it is shown on Table \ref{tab:action_vel}. 
The network structure of the DQN and DDQN is shown in Fig \ref{fig:architecture}. 

\begin{table}
\caption{Output actions and equivalent angular velocity}
\begin{center}
\begin{tabular}{r@{\quad}rl}
\hline
\multicolumn{1}{l}{\rule{0pt}{12pt}
                   Output}&\multicolumn{2}{l}{Angular velocity}\\[2pt]
\hline\rule{0pt}{12pt}
    0 & $-1.5$ $rad/s$  \\
    1 & $-0.75$ $rad/s$ \\
    2 & $0$ $rad/s$ \\
    3 & $0.75$ $rad/s$ \\
    4 & $1.5$ $rad/s$  \\[2pt]
\hline
\label{tab:action_vel}
\end{tabular}
\end{center}
\end{table}

The actor-network has as input the current state of the mobile robot followed by three fully-connected neural network layers with 256 nodes.
This input of the networks is transformed on the angular velocity that will be the commands sent to the motor of the mobile robot. 
The linear velocity is fixed and set on 0.15m/s, and there's no backward move.

\subsection{Reward Function}
Once the simulated environments are defined, it is possible to simulate the robot in the task of navigation.
First, the Deep-RL network needs to have the reward and penalty mechanism specified.
The rewards and penalties given to the agent are just numbers of a function that models how it wants the agent to behave. It can be based on empirical knowledge and created during the process of a solution to the problem.

Regarding the reward system, only three rewards were given, one for making the task correctly, the second in case of failure and the last in case of idle. The agent receives $200$ of reward when the goal is reached in a margin of $0.25$ meters. 
In a collision case against an obstacle or reaching the scenario limits, a negative reward of $-20$ is given. The collisions are verified if the distance sensor readings are lower than a distance of $0.12$ meters. Meanwhile, if the agent stays away from the target and  doesn't collide in a range of 500 steps, it's considered a condition of idle. In this case, the episode ends and a reward of $0$ is given.
This simplified rewarding system also helps focus on the Deep-RL approaches, their similarities, and differences, instead of the scenario.

\section{Results}\label{results}

\begin{figure*}
  \subfloat[First scenario.\label{fig:reward_1}]{
	\begin{minipage}[c][0.75\width]{0.32\textwidth}
	   \centering
	   \includegraphics[width=\textwidth]{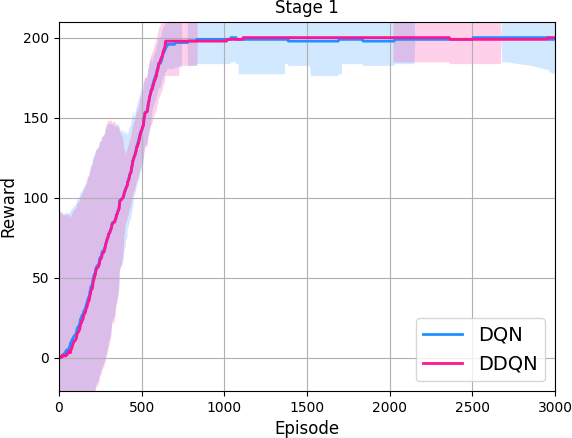}
	\end{minipage}}
 \hfill 	
  \subfloat[Second scenario.\label{fig:reward_2}]{
	\begin{minipage}[c][0.75\width]{0.32\textwidth}
	   \centering
	   \includegraphics[width=\textwidth]{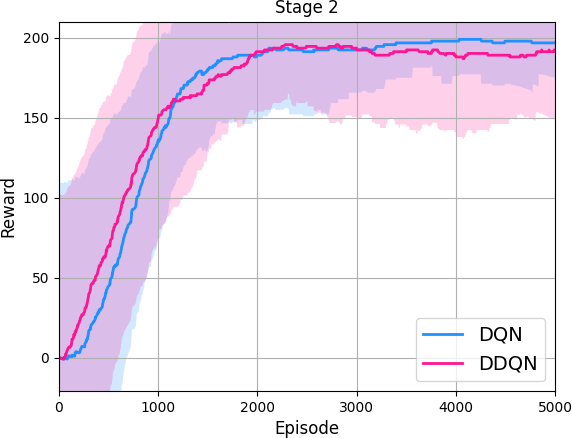}
	\end{minipage}}
 \hfill 
 \subfloat[Third scenario.\label{fig:reward_3}]{
	\begin{minipage}[c][0.75\width]{0.32\textwidth}
	   \centering
	   \includegraphics[width=\textwidth]{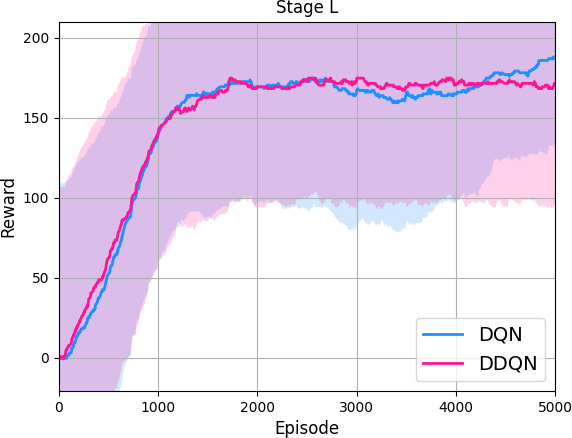}
	\end{minipage}}
\caption{Moving average of the reward over $3000$, $5000$, and $5000$ training episodes for the respectively scenarios.}
\label{fig:rewards}
\end{figure*}

\begin{figure*}
  \subfloat[DQN\label{fig:dqn_st1}]{
	\begin{minipage}[c][\width]{0.16\textwidth}
	   \centering
	   \includegraphics[width=\textwidth]{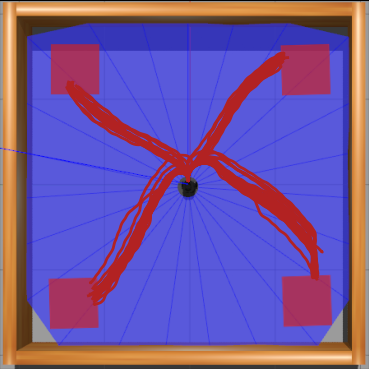}
	\end{minipage}}
  \subfloat[DDQN\label{fig:ddqn_st1}]{
	\begin{minipage}[c][\width]{0.16\textwidth}
	   \centering
	   \includegraphics[width=\textwidth]{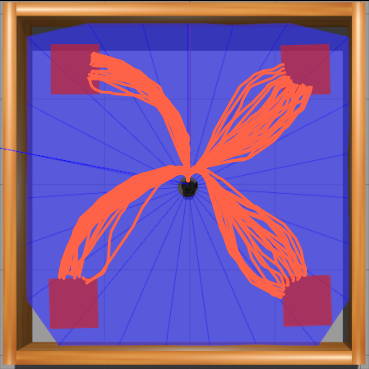}
	\end{minipage}}
 \subfloat[DQN\label{fig:dqn_st2}]{
	\begin{minipage}[c][\width]{0.16\textwidth}
	   \centering
	   \includegraphics[width=\textwidth]{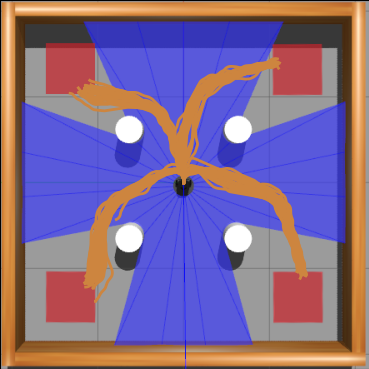}
	\end{minipage}}
  \subfloat[DDQN\label{fig:ddqn_st2}]{
	\begin{minipage}[c][\width]{0.16\textwidth}
	   \centering
	   \includegraphics[width=\textwidth]{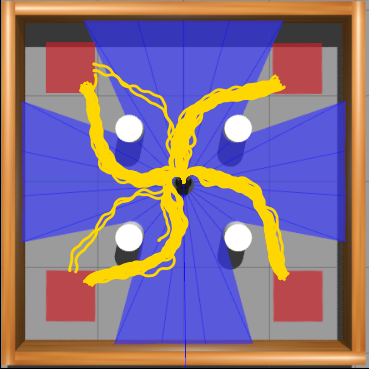}
	\end{minipage}}
  \subfloat[DQN\label{fig:dqn_st3}]{
	\begin{minipage}[c][\width]{0.16\textwidth}
	   \centering
	   \includegraphics[width=\textwidth]{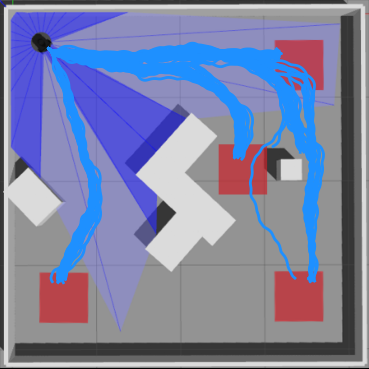}
	\end{minipage}}
  \subfloat[DDQN\label{fig:ddqn_st3}]{
	\begin{minipage}[c][\width]{0.16\textwidth}
	   \centering
	   \includegraphics[width=\textwidth]{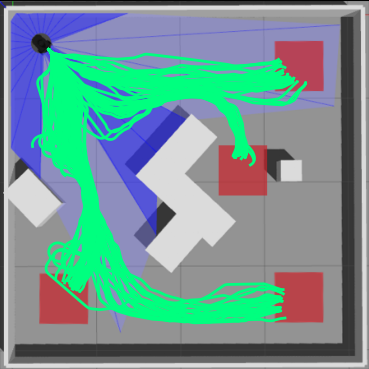}
	\end{minipage}}
\caption{The behavior of each approach was tested in each simulated scenario over
100 navigation trials.}
\label{fig:behavior_simulation}
\end{figure*}

This section highlights the results yielded in this work. For each scenario and model, an extensive amount of statistics are collected. The evaluation was done in a simulation workplace, and the test tasks were performed for 100 trials in each pre trained model. Considering there are 4 fixed goals, it is performed 25 trials for each fixed goal, and the total of successful trials was recorded. Also, the average navigation time with their standard deviations is recorded. The learning through time can be seen in Figure \ref{fig:rewards}, while the behavior of the agents during the evaluation can be seen in Figure \ref{fig:behavior_simulation}. Also, Table \ref{table:mean_std} shows the overall results gathered.

Overall, one of the main contributions of this work can be observed in Figure \ref{fig:rewards}. We can see how the learning converged stably throughout the three scenarios that were used to evaluate our agents. From the results of many related works, it is possible to see that the agents manage to achieve the best rewards possible in a fast manner; and, most importantly, showing to be stable and consistent after achieving the learning. In this high complexity scenario - mainly in the third scenario - it would be normal if the agent presented some kind of forgetting or even forgetting to do it at all. However, it can be observed that the agent managed to maintain stable reward throughout the evolution. 

These contributions can be validated by the statistics of Table \ref{table:mean_std}, where it is possible to see that both agents managed to achieve a good performance. The overall performance, with almost 100\% precision for all algorithms and scenarios, was reached by the Double Q algorithm, which is very suitable for performing continuous motion in terrestrial mobile robots. Even though these algorithms are not as sophisticated as the constrastive learning algorithms or the other more recent Deep-RL algorithms for continuous actions, we can see that the simplicity of the methodology presented is capable of reaching a good performance.

\begin{table}
\caption{Mean and standard deviation metrics over $100$ navigation trials in three different simulated scenarios for all approaches. The Episode Time ($ET$) and Success Rate ($SR$) are presented for each test.}
\begin{center}
\begin{tabular}{cccc}
\hline
\multicolumn{1}{c}{\rule{0pt}{12pt} Env} &  {Algorithm} &  {${ET}_{sim}$} & {${SR}_{sim}$}\\[2pt]
\hline\rule{0pt}{12pt}
1 & DQN & $16.02$ $\pm$ $1.67$ & $100$\% \\
1 & DDQN & $15.72$ $\pm$ $1.06$ & $100$\% \\
2 & DQN & $17.13$ $\pm$ $0.28$ & $100$\% \\
2 & DDQN & $17.39$ $\pm$ $1.15$ & $100$\% \\
3 & DQN & $26.05$ $\pm$ $6.94$ & $89$\%  \\
3 & DDQN & $25.60$ $\pm$ $7.73$ & $97$\%  \\ [2pt]
\hline
\label{table:mean_std}
\end{tabular}
\end{center}
\end{table}

\section{Conclusions}\label{conclusions}

In this paper, we presented two simple Deep-RL approaches that rely on low-dimensional data to achieve a good performance on motion tasks for terrestrial mobile robots. It was shown that Double Q algorithms are capable of providing good performance even in relation to sophisticated Deep-RL algorithms, such as actor-critic or constrastive architectures. This validation was achieved by comparing the testing results in the simulated environments. 

Another interesting conclusion was the stable pattern of the learning process shown throughout different scenarios and the time, showing small signs of typical Deep-RL problems like gradient convergence or the forgetting. More studies are on the way to confirm this pattern through other types of mobile robots and Deep-RL techniques.

%
%


\begin{thebibliography}{6}
%

\bibitem {kober2013reinforcement}
Kober, J., Bagnell, J.A., Peters, J.: Reinforcement learning in robotics: A survey.
The International Journal of Robotics Research 32(11), 1238–1274 (2013)

\bibitem {laskin2020curl}
Laskin, M., Srinivas, A., Abbeel, P.: Curl: Contrastive unsupervised representations
for reinforcement learning. In: International Conference on Machine Learning. pp.
5639–5650. PMLR (2020)

\bibitem {tai2017virtual}
Tai, L., Paolo, G., Liu, M.: Virtual-to-real deep reinforcement learning: Continuous
control of mobile robots for mapless navigation. In: Intelligent Robots and Systems
(IROS), 2017 IEEE/RSJ International Conference on. pp. 31–36. IEEE (2017)

\bibitem {jesus2021soft}
Jesus, J.C., Kich, V.A., Kolling, A.H., Grando, R.B., Cuadros, M.A.d.S.L.,
Gamarra, D.F.T.: Soft actor-critic for navigation of mobile robots. Journal of In-
telligent \& Robotic Systems 102(2), 1–11 (2021)

\bibitem {grando2022double}
Grando, R.B., de Jesus, J.C., Kich, V.A., Kolling, A.H., Drews-Jr, P.L.J.: Double
critic deep reinforcement learning for mapless 3d navigation of unmanned aerial
vehicles. Journal of Intelligent \& Robotic Systems 104(2), 1–14 (2022)

\bibitem {carlucho2018}
Carlucho, I., De Paula, M., Wang, S., Menna, B.V., Petillot, Y.R., Acosta, G.G.:
Auv position tracking control using end-to-end deep reinforcement learning. In:
MTS/IEEE OCEANS (2018).

\bibitem {bedin2021deep}
Grando, R.B., de Jesus, J.C., Kich, V.A., Kolling, A.H., Bortoluzzi, N.P., Pinheiro,
P.M., Alves Neto, A., Drews-Jr, P.L.J.: Deep reinforcement learning for mapless
navigation of a hybrid aerial underwater vehicle with medium transition. In: IEEE
ICRA. pp. 1088–1094 (2021).

\bibitem {chen2017socially}
Chen, Y.F., Everett, M., Liu, M., How, J.P.: Socially aware motion planning with
deep reinforcement learning. In: Intelligent Robots and Systems (IROS), 2017
IEEE/RSJ on. pp. 1343–1350.

\bibitem {jesus2019deep}
Jesus, J.C., Bottega, J.A., Cuadros, M.A., Gamarra, D.F.: Deep deterministic pol-
icy gradient for navigation of mobile robots in simulated environments. In: 2019
International Conference on Advanced Robotics (ICAR). pp. 362–367. IEEE
(2019)

\bibitem {mnih2013playing}
Mnih, V., Kavukcuoglu, K., Silver, D., Graves, A., Antonoglou, I., Wierstra, D.,
Riedmiller, M.: Playing atari with deep reinforcement learning. arXiv preprint
arXiv:1312.5602 (2013)

\bibitem {hausknecht2015deep}
Hausknecht, M., Stone, P.: Deep recurrent q-learning for partially observable mdps.
In: 2015 AAAI Fall Symposium Series (2015)

\bibitem {van2016deep}
Van Hasselt, H., Guez, A., Silver, D.: Deep reinforcement learning with double
q-learning. In: Thirtieth AAAI Conference on Artificial Intelligence (2016)

\bibitem {lillicrap2015continuous}
Lillicrap, T.P., Hunt, J.J., Pritzel, A., Heess, N., Erez, T., Tassa, Y., Silver, D.,
Wierstra, D.: Continuous control with deep reinforcement learning. (2015)

\bibitem {dobrevski2018map}
Dobrevski, M., Skocaj, D.: Map-less goal-driven navigation based on reinforcement
learning. In: 23rd Computer Vision Winter Workshop (2018)

\bibitem {tai2016towards}
Tai, L., Liu, M.: Towards cognitive exploration through deep reinforcement learning
for mobile robots. CoRR abs/1610.01733 (2016)

\bibitem {hasselt2010double}
Hasselt, H.: Double q-learning. Advances in neural information processing systems
23 (2010)

\bibitem {Pfitscher2019}
Pfitscher, M., Welfer, D., do Nascimento, E.J., Cuadros, M.A.d.S.L., Gamarra,
D.F.T.: Article users activity gesture recognition on kinect sensor using convolu-
tional neural networks and fastdtw for controlling movements of a mobile robot.
Inteligencia Artificial 22(63), 121–134 (Apr 2019)

\bibitem {daSilva2019}
da Silva, R.M., de Souza Leite Cuadros, M.A., Gamarra, D.F.T.: Comparison of a
backstepping and a fuzzy controller for tracking a trajectory with a mobile robot.
In:  Intelligent Systems Design and Applications. pp. 212–221. Springer International Publishing, Cham (2020)

\bibitem {subramanian2018deep}
Subramanian, V.: Deep Learning with PyTorch: A practical approach to building
neural network models using PyTorch. Packt Publishing Ltd (2018)

\bibitem {pyo2015ros}
Pyo, Y., Cho, H., Jung, R., Lim, T.: Ros robot programming. Seoul, ROBOTIS
Co (2015)



\bibitem {fairchild2016ros}
Fairchild, C., Harman, T.L.: ROS Robotics By Example. Packt Publishing Ltd
(2016)















\end{thebibliography}
\end{document}